\documentclass[10pt,twocolumn,letterpaper]{article}

\usepackage{cvpr}              
\usepackage[table]{xcolor}
\definecolor{cvprblue}{rgb}{0.21,0.49,0.74}

\definecolor{gold}{rgb}{0.890,0.831,0.714}
\definecolor{silver}{rgb}{0.21,0.49,0.74}
\definecolor{bronze}{rgb}{0.21,0.49,0.74}
\usepackage[pagebackref,breaklinks,colorlinks,allcolors=cvprblue]{hyperref}

\def\paperID{*****} 
\def\confName{CVPR}
\def\confYear{2026}

\title{NTIRE 2026 Low-light Enhancement: Twilight Cowboy Challenge}

\author{
Aleksei Khalin
\and
Egor Ershov
\and
Artyom Panshin
\and
Sergey Korchagin
\and
Georgiy Lobarev
\and
Arseniy Terekhin
\and
Sofiia Dorogova
\and
Amir Shamsutdinov
\and
Yasin Mamedov
\and
Bakhtiyar Khalfin
\and
Bogdan Sheludko
\and
Emil Zilyaev
\and
Nikola Bani\'c
\and
Georgy Perevozchikov
\and
Radu Timofte
\and
Shuai Liu
\and
Yuqian Zhang
\and
Lize Zhang
\and
Yibin Huang
\and
Chaoyu Feng
\and
Luyang Wang
\and
Xiaotao Wang
\and
Dongqing Zou
\and
Lei Lei
\and
Tianli Liu
\and
Dejun Hao
\and
Chunxia Lei
\and
Furkan K\i nl\i
\and
Andrei Mironov
\and
Alexander Dikov
\and
Aleksei Sadokhin
\and
Vladimir Zvorygin
\and
Constantine Habarlak
\and
Shuwei Yue
\and
Egor Mirantsov
\and
Daniil Okunev
\and
Dmitry Arkhipov
\and
Aleksandr Yugay
\and
Anas M. Ali
\and
Bilel Benjdira
\and
Wadii Boulila
\and
Wei Zhou
\and
Linfeng Li
\and
Lingdong Kong
\and
Jiachen Tu
\and
Guoyi Xu
\and
Yaoxin Jiang
\and
Jiajia Liu
\and
Yaokun Shi
}

\begin{document}
\maketitle
    
\begin{abstract}

This paper presents a review of the NTIRE 2026 Low-light Enhancement: Twilight Cowboy Challenge. 
The objective of the competition was to merge a set of misaligned smartphone images in the raw domain, captured in low-light conditions, into a single, clean image. 
Introduced setup simultaneously addresses two problems of low-light photography: visual degradations such as high noise and mixed scene illuminants, and the geometric inconsistencies caused by hand movement during multi-frame capture.
To advance research in low-light and nighttime computational photography, a challenging dataset was collected comprising 585 real-world scenes, spanning indoor low-light and outdoor nighttime conditions, for training and benchmarking participant solutions.
The competition employed a three-stage evaluation protocol: automatic validation via the CodaBench platform in stages one and two, followed by blind assessment on a private test set for the final ranking. 
Ten teams surpassed the established baseline, achieving improvements of up to +6.49 dB in PSNR and +0.0101 in SSIM, thereby establishing new state-of-the-art performance for burst-based low-light image enhancement.
These results demonstrate significant progress in handling real-world noise, motion, and illumination variability in the low-light setting.
Comprehensive results, leaderboards, and additional information are publicly available at \url{https://nightimaging.org}.


\end{abstract}

\section{Introduction}
\label{sec:intro}

\begin{figure}[t]
\centering
\includegraphics[width=\linewidth]{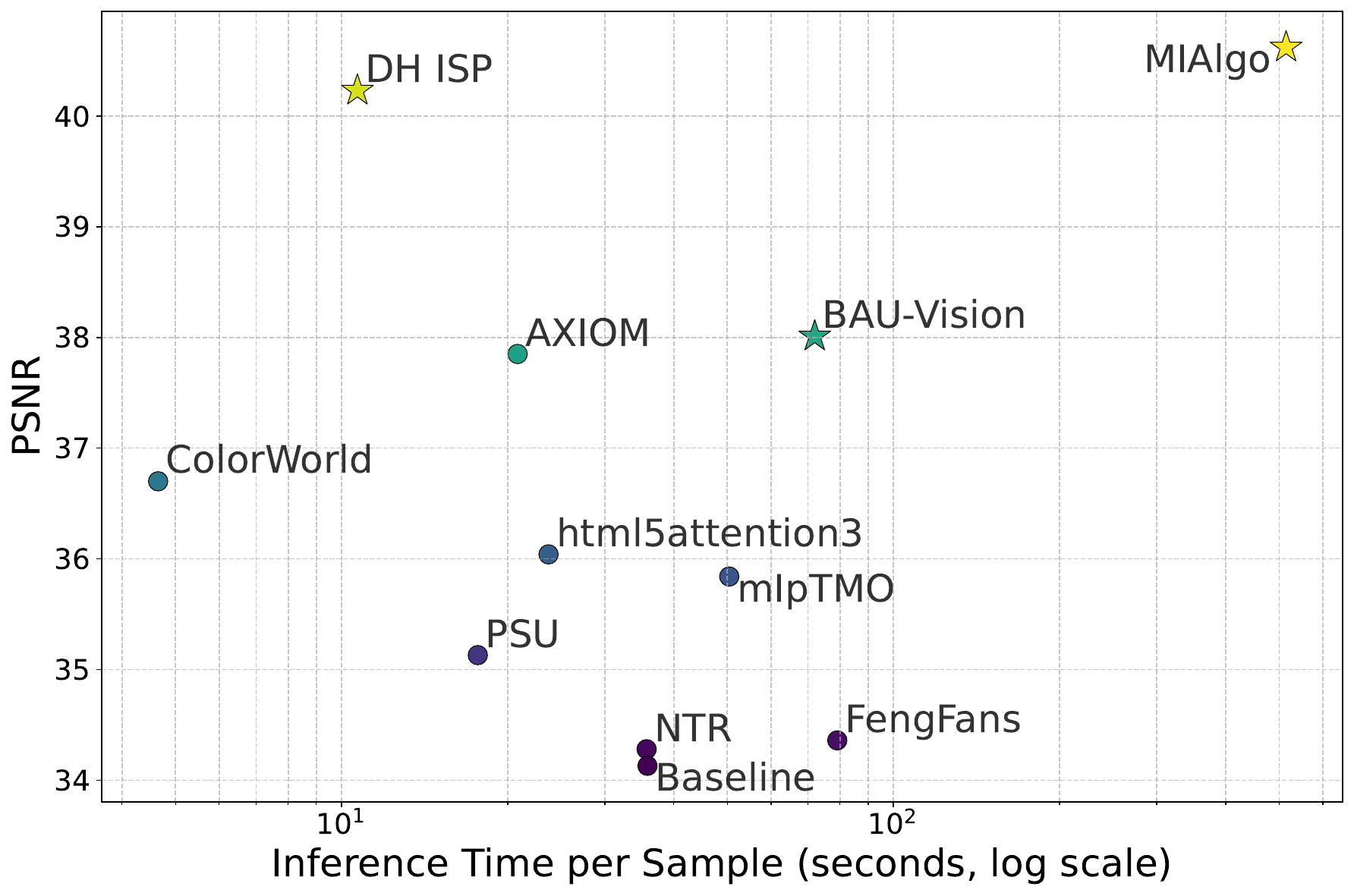}
\caption{Comparison of mean PSNR scores and inference times per sample for participant solutions. Top-3 solutions are displayed with star markers.}
\label{fig:performance}
\end{figure}

Image denoising is a fundamental component of image signal processing. 
Its primary objective is to suppress noise while preserving fine details and enhancing overall image quality. 
Among various scenarios, low-light imaging presents one of the most challenging conditions for denoising due to inherently low signal levels. 
A straightforward solution is to increase the exposure time; however, while this approach effectively reduces noise, it also makes the image more susceptible to motion blur caused by camera or scene movement. 
In practical situations, where users often lack stabilization tools such as tripods, this trade-off typically results in images that are either noisy or blurred.

Smartphones, however, can employ a technique known as burst imaging. 
In this approach, a sequence of short-exposure images is captured in rapid succession, resulting in a set of sharp yet noisy observations of the same scene. 
Owing to the approximately unbiased nature of noise, such frames can be effectively aggregated --- provided accurate alignment --- to approximate the quality of a single long-exposure image. 
Within this framework, the problem of enhancing low-signal data can be naturally decomposed into two key sub-tasks: image alignment and image aggregation.

We propose a challenge focused on generating a denoised image from a collection of misaligned low-exposure frames. 
With this motivation, the primary objective of the challenge is to further stimulate research in advanced image processing techniques for night photography. 

This challenge is one of the challenges associated with the NTIRE 2026 Workshop\footnote{\url{https://www.cvlai.net/ntire/2026/}} on:
deepfake detection~\cite{ntire26deepfake}, 
high-resolution depth~\cite{ntire26hrdepth},
multi-exposure image fusion~\cite{ntire26raim_fusion}, 
AI flash portrait~\cite{ntire26raim_portrait}, 
professional image quality assessment~\cite{ntire26raim_piqa},
light field super-resolution~\cite{ntire26lightsr},
3D content super-resolution~\cite{ntire263dsr},
bitstream-corrupted video restoration~\cite{ntire26videores},
X-AIGC quality assessment~\cite{ntire26XAIGCqa},
shadow removal~\cite{ntire26shadow},
ambient lighting normalization~\cite{ntire26lightnorm},
controllable Bokeh rendering~\cite{ntire26bokeh},
rip current detection and segmentation~\cite{ntire26ripdetseg},
low light image enhancement~\cite{ntire26llie},
high FPS video frame interpolation~\cite{ntire26highfps},
night-time dehazing~\cite{ntire26nthaze,ntire26nthaze_rep},
learned ISP with unpaired data~\cite{ntire26isp},
short-form UGC video restoration~\cite{ntire26ugcvideo},
raindrop removal for dual-focused images~\cite{ntire26dual_focus},
image super-resolution (x4)~\cite{ntire26srx4},
photography retouching transfer~\cite{ntire26retouching},
mobile real-word super-resolution~\cite{ntire26rwsr},
remote sensing infrared super-resolution~\cite{ntire26rsirsr},
AI-Generated image detection~\cite{ntire26aigendet},
cross-domain few-shot object detection~\cite{ntire26cdfsod},
financial receipt restoration and reasoning~\cite{ntire26finrec},
real-world face restoration~\cite{ntire26faceres},
reflection removal~\cite{ntire26reflection},
anomaly detection of face enhancement~\cite{ntire26anomalydet},
video saliency prediction~\cite{ntire26videosal},
efficient super-resolution~\cite{ntire26effsr},
3D restoration and reconstruction in adverse conditions~\cite{ntire26realx3d},
image denoising~\cite{ntire26denoising},
blind computational aberration correction~\cite{ntire26aberration},
event-based image deblurring~\cite{ntire26eventblurr},
efficient burst HDR and restoration~\cite{ntire26bursthdr},
and efficient low light image enhancement~\cite{ntire26effllie}.

\section{Twilight Cowboy Challenge overview}
\label{sec:overview}


Nighttime image processing presents numerous challenges, primarily due to the presence of strong light sources within the scene and high noise levels. 
One effective approach to mitigating these issues is burst capture, which significantly reduces noise by combining multiple frames of the same scene. 
However, this technique introduces its own difficulties -- handheld shooting inevitably involves slight camera motion, complicating frame alignment and registration.

In this competition, participants are tasked with developing a method to transform a set of unprocessed RAW images captured in a handheld setting into a a denoised RAW image equivalent to one captured with a stationary camera. 
This task introduces several novel aspects that distinguish it from existing approaches in computational photography. 
It requires joint motion compensation and denoising directly in the Bayer/RAW domain -- a significantly more complex inverse problem than conventional pipelines that operate on already demosaiced RGB images or assume perfect alignment. 
By explicitly requiring robustness to natural hand tremor, the challenge bridges the gap between idealized laboratory conditions and practical mobile photography scenarios. 
These aspects position the challenge at the forefront of computational photography research, addressing a critical yet underexplored gap in bringing professional-grade low-light imaging capabilities to everyday handheld devices.


\subsection{Data collection}

The training data consist of pairs, each containing one RAW ground-truth image and five RAW images captured with misaligned RAW images of the same low-light scene captured under handheld conditions. 
The dataset includes both indoor and outdoor scenes. 

First, the smartphone was mounted on a tripod, and 270 frames were captured in RAW format using a remote trigger to eliminate camera shake. 
These frames were later used to construct the ground-truth image, as detailed in a subsequent section. 
To collect training samples, the capture initially began with the phone on the tripod, ensuring alignment with the static ground-truth image. 
The phone was then handheld to introduce natural motion. 
This motion reflects the conditions of burst photography. 
From this handheld sequence, every tenth frame was selected, yielding five images with varying levels of misalignment due to camera movement.

All images were collected with fixed camera settings: ISO 100, shutter speed 1/10 s, aperture f/1.8, and focus distance set to $\infty$. 

\subsection{Ground truth generation}
To produce high-quality, noise-free images from captured data, we adopt practices established in the Smartphone Image Denoising Dataset~\cite{SIDD_2018_CVPR}.
The 270 frames captured for each scene first underwent intensity alignment: the average intensity was computed for each frame, and a normal distribution was fitted to these values of the scene.
Frames falling outside the 99.7\% interval were discarded.
The most common cause of intensity misalignment in nighttime scenes was movement of objects outside the field of view, e.g. moving cars with lights.
The remaining frames were then intensity-aligned with the estimated mean intensity of the sequence.

Next, robust mean image estimation was performed at the pixel level.
To do so, the a normal distribution was fitted to the corresponding values from the remaining frames, and the mean of the distribution was taken as the ground-truth value.

Finally, to improve visual appearance, quantile-based normalization was used. For each image $I$, the final image $\hat I$ is given by
\begin{equation}
    \hat I = \frac{I - \mathcal{Q}(I, 0.01)}{\mathcal{Q}(I, 0.99) - \mathcal{Q}(I, 0.01)}, \label{eq:quantile-norm}
\end{equation}
where $\mathcal{Q}(I, k)$ denotes the $k^{\mathrm{th}}$ quantile of pixel intensities in image $I$.

\subsection{Data structure}

The five burst photography images and the ground truth estimate obtained in a previous step were used as input and reference, respectively. 
RAW images were also accompanied with the metadata stored by the device.

The challenge was structured in three stages.
The initial stage participants were provided with 250 training pairs and 50 validation samples (without ground truth) for algorithm development.
In the second stage, an additional 150 training pairs and 50 validation samples were released.
The final testing dataset consisted of 85 input–ground-truth pairs and was not released to participants during the challenge.
Approximately one third of the test data comprises indoor scenes. 
Additionally, baseline code was provided as a starting point.

\subsection{Evaluation and placement}

All submitted solutions were evaluated on the test subset of the collected dataset.
To evaluate the solutions, mean PSNR~\cite{gonzalez2018digital} and SSIM~\cite{wang2004image} values were calculated across all testing images.
The participants were then ranked according to mean PSNR and mean SSIM separately, resulting in the PSNR rank $R_{\mathrm{P}}$ and SSIM rank $R_{\mathrm{S}}$ values for each team.
The final score was computed as
\begin{equation}
    S = 0.6\cdot R_{\mathrm{P}} + 0.4 \cdot R_{\mathrm{S}}, \label{eq:final-score}
\end{equation}
prioritizing solutions with better PSNR values over those with higher SSIM.
Teams were ranked in ascending order of the final score $S$ for the final leaderboard.
\section{Results}
\label{sec:results}


The section presents the evaluation results and descriptions of all submitted solutions, along with the baseline. 
Ranking results are presented in Table~\ref{tab:results}, together with the inference time per sample.
Inference time did not affect final placement, but it provides insight into image processing efficiency and relationship between reconstruction quality and model complexity.

\begin{table}[ht]
\centering
\begin{tabular}{lrrrr}
\toprule
Team & PSNR & SSIM & $S$ & $T_{\mathrm{avg}}$, s \\
\midrule
MiAlgo & 40.62 & 0.9875 & 1.0 & 514.6 \\
DH ISP & 40.23 & 0.9873 & 2.0 & 10.7 \\
BAU-Vision & 38.01 & 0.9855 & 3.4 & 72.4 \\
AXIOM & 37.85 & 0.9865 & 3.6 & 20.9 \\
html5attention3 & 36.04 & 0.9829 & 5.6 & 23.7 \\
ColorWorld & 36.70 & 0.9777 & 6.6 & 4.7 \\
mIpTMO & 35.84 & 0.9815 & 7.0 & 50.4 \\
PSU & 35.13 & 0.9822 & 7.2 & 17.7 \\
FengFans & 34.36 & 0.9814 & 8.6 & 79.1 \\
NTR & 34.28 & 0.9755 & 10.4 & 35.7 \\
\textit{Baseline} & 34.13 & 0.9774 & 10.6 & 35.9 \\
\bottomrule
\end{tabular}
\caption{Final leaderboard of solutions on the testing dataset, with mean inference time per sample, denoted by $T_{\mathrm{avg}}$. Participants are ranked by their final score $S$, given by Eq.~\ref{eq:final-score}, in ascending order.}
\label{tab:results}
\end{table}

\begin{figure*}[t]
\centering
\includegraphics[width=0.75\textwidth]{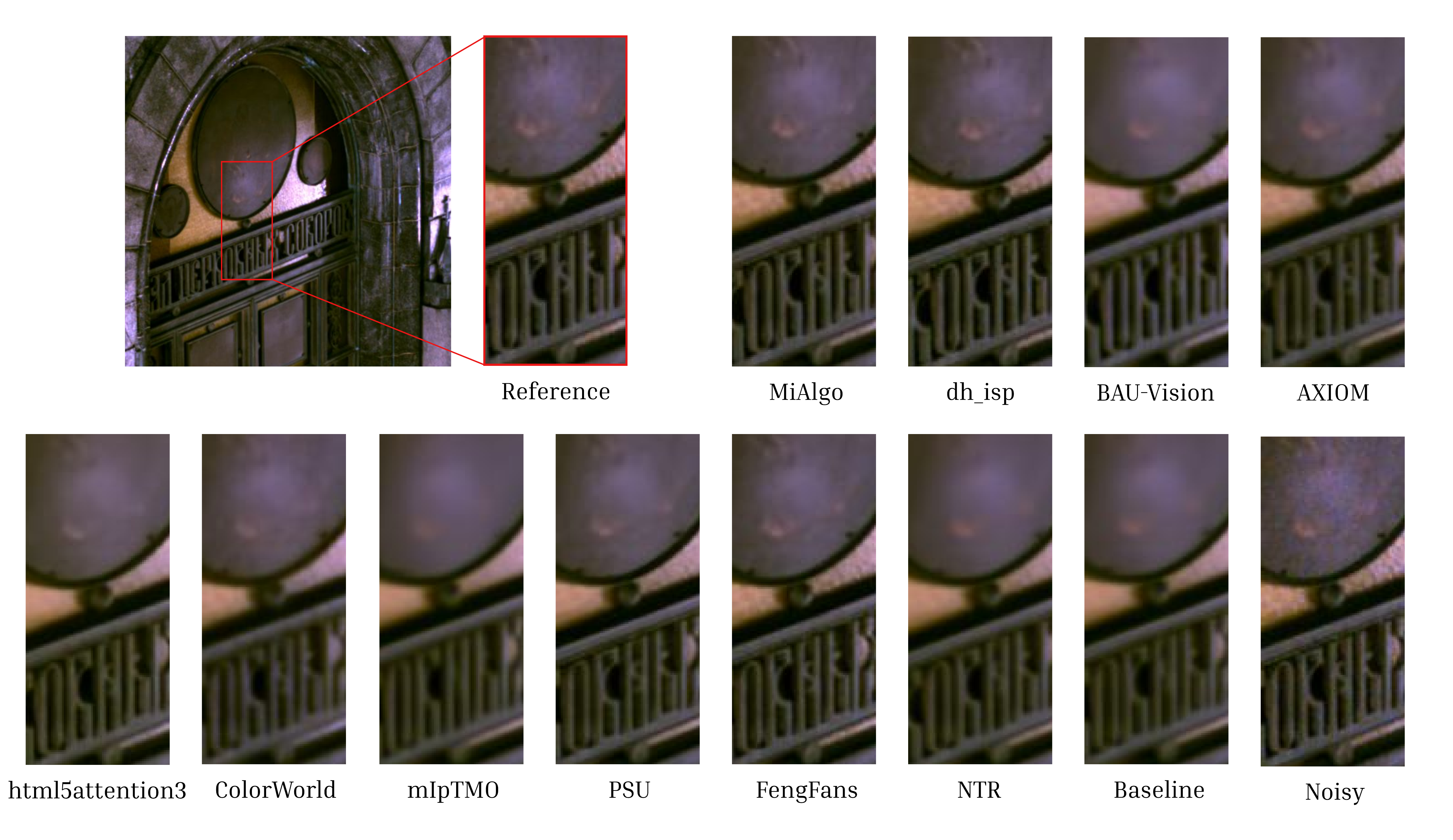}
\caption{Visual comparison of denoising approaches submitted to the final stage of the competition. Top-performing solutions excel at preserving fine details while effectively removing noise.}
\label{fig:solution-comp}
\end{figure*}

Out of the submitted solutions, the best results were demonstrated by teams MiAlgo, DH ISP, and BAU-Vision ranked first through third, respectively.
The solution presented by team ColorWorld is the most lightweight, requiring only 4.7 s per sample and maintaining good reconstruction quality. 
Team DH ISP offers the best balance between performance and efficiency, with an inference time of only 10.7 s per sample with a high PSNR of 40.23 dB.
Finally, the model of team MiAlgo demonstrates the performance achievable with high computational complexity, reaching 40.62 dB PSNR and an SSIM value of 0.9875, effectively representing the current state-of-the-art.
Fig.~\ref{fig:performance} presents a comparison of PSNR values and inference times across all solutions, and Fig.~\ref{fig:solution-comp} showcases the outputs for a scene with complex textures and fine details.
Below are descriptions of the solutions presented by their respective authors.

\subsection{Baseline}

In this challenge, simple alignment of frames with averaging and BM3D~\cite{dabov2007bm3d} denoising was given as a baseline method to participants. 

For alignment we employ detector-free matching with pretrained ASpanFormer \cite{chen2022aspanformer}. 
As a preliminary step, RAW images were rendered to sRGB before being fed into ASpanFormer, mitigating potential domain gap issues.
The predicted keypoint correspondences are then used with the RANSAC algorithm for robust homography estimation.
The final step of multi-frame aggregation is applying homographies to align images to the reference frame and averaging all 5 frames.

The resulting averaged frame, with reduced noise, was subsequently rocessed using the BM3D algorithm applied independently to each RAW channel. The noise parameter $\sigma$ is selected to maximize PSNR.
After selecting $\sigma = 3/2047$, it is fixed for all images from the assumption that all images have approximately the same noise level.

The final output image is obtained using robust percentile-based normalization, where the RAW intensities are linearly scaled using the 1st and 99th percentiles, following Eq.~\ref{eq:quantile-norm}.


\subsection{MiAlgo}

\begin{figure}[t]
\centering
\includegraphics[width=\linewidth]{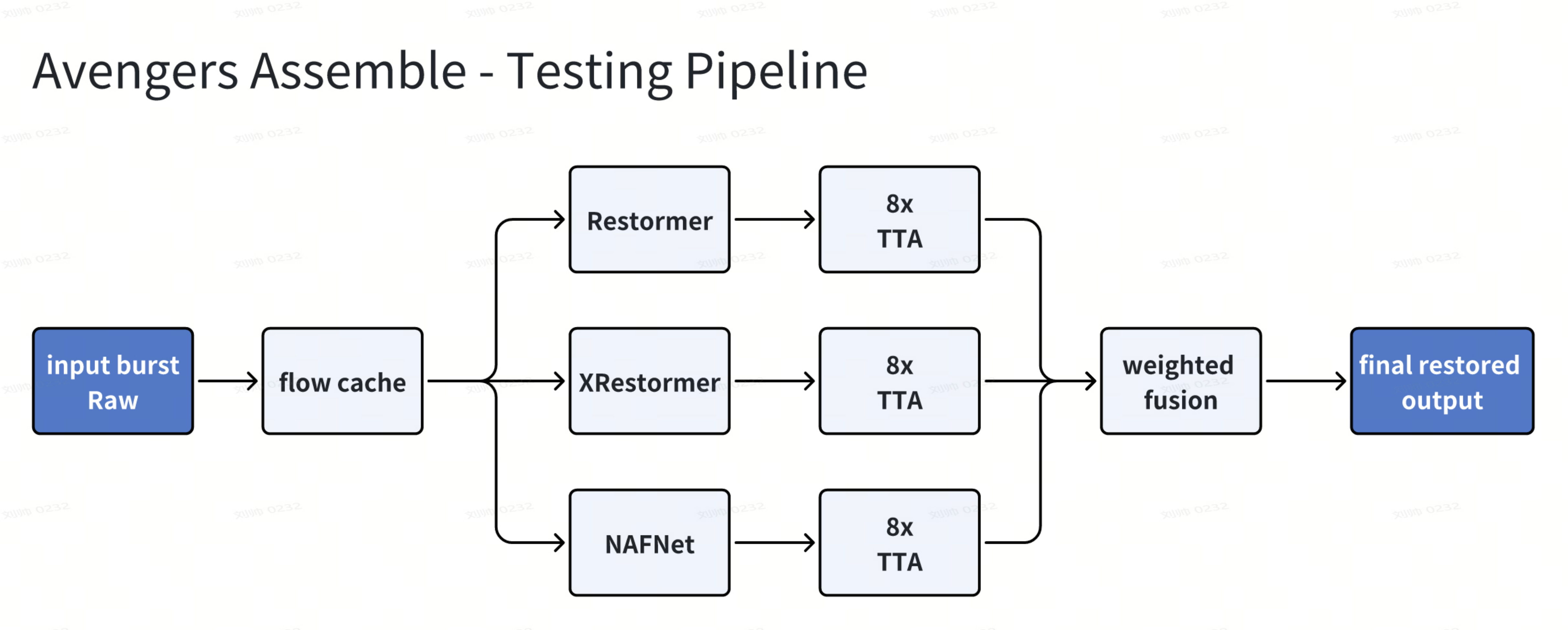}
\caption{Testing pipeline of Avengers Assemble (MiAlgo).}
\label{fig:MiAlgo_pipeline}
\end{figure}

\textbf{Dual-ensemble framework.} Our method is built on a dual-ensemble design that combines model ensembling and test-time augmentation (TTA) ensembling. We first run three complementary restoration models and fuse their predictions with weighted averaging, then enhance robustness by averaging multiple geometric TTA outputs per model. This two-level aggregation improves stability across diverse night scenes, especially when a single model may fail on specific texture, noise, or motion patterns.

\textbf{Flow-guided deformable alignment.} To improve temporal correspondence in burst inputs, we precompute optical flow and inject flow cues into deformable alignment modules (flow-guided deformable convolution style). This design is inspired by recent burst HDR/restoration methods~\cite{rmea_spatiotemporal_decoupling,flow_guided_deformable_alignment_reconstruct} and helps alignment modules estimate offsets under large inter-frame motion. Compared with purely image-driven alignment, flow-guided alignment better preserves local structures and reduces motion-induced artifacts in dynamic regions.

\textbf{Sequential multi-model denoising ensemble.} The denoising stage consists of three mainstream backbones: Restormer~\cite{Zamir22}, XRestormer, and NAFNet~\cite{chen2022simple}. We execute them sequentially and accumulate their weighted outputs into a final prediction. The three models offer complementary inductive biases: transformer-style global modeling (Restormer/XRestormer) and efficient convolutional restoration behavior (NAFNet). Their combination improves overall reconstruction robustness and detail fidelity in low-light RAW restoration.

\textbf{TTA ensemble with flow-image consistency.} For each model, we apply 8-way geometric TTA. RAW frames and flow maps are transformed consistently before inference, and outputs are inversely transformed and averaged afterward. This consistency constraint is important: if image and flow transforms are mismatched, alignment quality degrades. The proposed TTA ensemble improves orientation robustness and reduces occasional prediction variance across challenging cases. Full testing pipeline can be seen in Figure \ref{fig:MiAlgo_pipeline}.

\textbf{Training strategy with manual data cleaning.} We use a two-stage training protocol. First, models are pretrained on the full training set to learn general restoration priors. Then we manually clean the dataset by removing samples with visible spatial misalignment between burst input and ground truth (about 25\% removed), and finetune on the cleaned subset. This cleaning step significantly improves spatial consistency and effectively reduces pixel-shift artifacts in final outputs.



\subsection{DH ISP}


\textbf{Data construction.} The workflow begins with noise residual extraction: coarsely aligning unregistered noisy-GT pairs to extract "noise residual maps" containing authentic sensor noise and minor motion blur. 
Subsequently, noise distribution modeling and transfer are performed by fitting the residuals using traditional statistical methods (PCA-based noise covariance analysis) and deep generative models (GAN-driven noise style transfer). 
The learned noise distribution is then precisely mapped back to GT images. This approach effectively eliminates training artifacts caused by misregistration, 
enabling the network to focus on learning noise statistics.

\begin{figure}[t]
  \centering
  \includegraphics[width=1.0\linewidth]{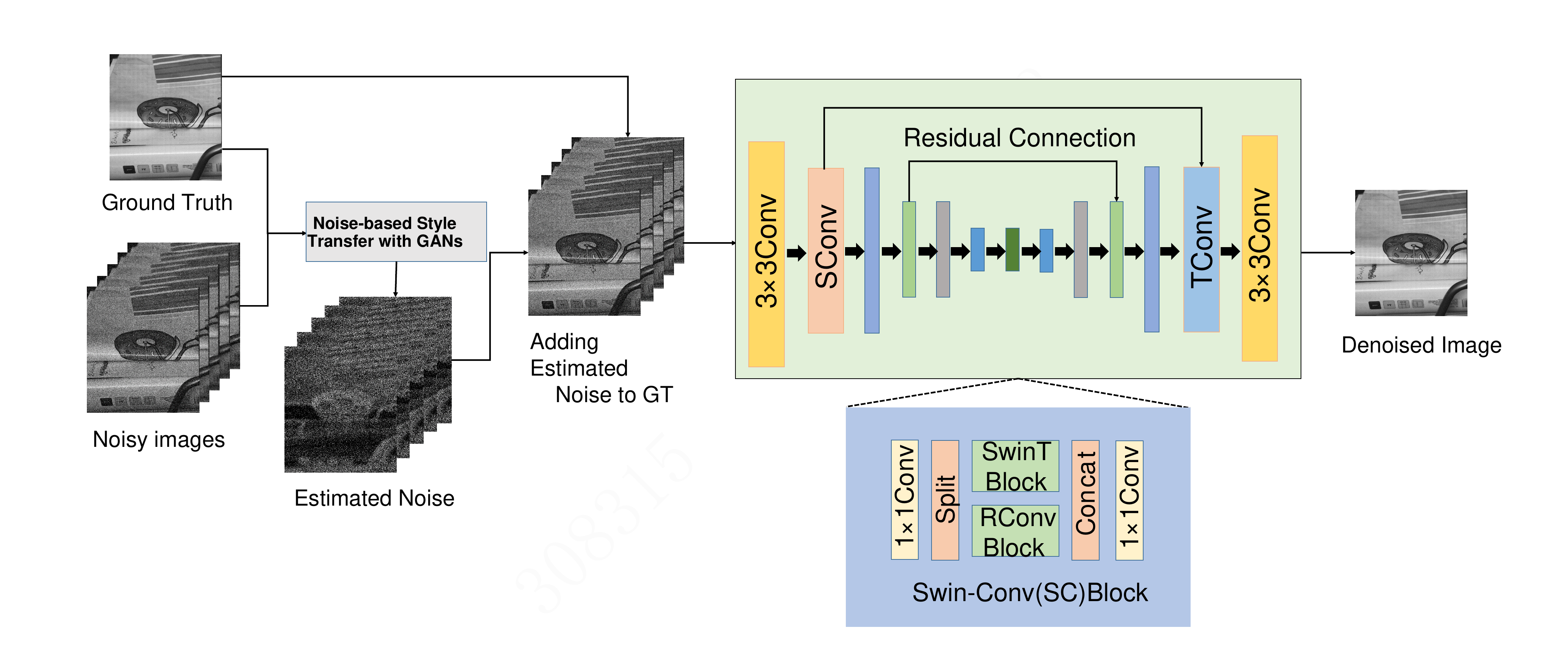}

  \caption{Overall Pipeline of ``High-Precision Noise Transfer and SCUNET-Based Multi-Frame RAW Image Denoising Solution'' (DH ISP).}
  \label{fig:onecol}
\end{figure}

\textbf{Implementation Details.} The framework combined ASpanFormer \cite{chen2022aspanformer} (with frozen weights) and SCUNET \cite{zhang2023practical} for multi-frame RAW denoising enhancement; 
the overall pipeline is illustrated in Figure~\ref{fig:onecol}. For SCUNET, we adopted a window size of 16 during training to achieve optimal performance. 
The denoising model was optimized using a composite loss combining L1 loss, perceptual loss, SSIM loss, and edge loss. 
\subsection{BAU-Vision}

Our methodology addresses high-fidelity restoration from RAW bursts by decoupling global luminance normalization and spatial registration from hierarchical feature restoration. The pipeline operates in the packed Bayer domain to preserve sensor statistics and mitigate interpolation-induced artifacts.

To stabilize restoration across heterogeneous exposures, we implement global brightness correction via a Gradient Boosting Regressor (GBR) \cite{friedman2001greedy}. The GBR is trained on a $53$-dimensional statistical vector, including intensity percentiles ($10^{th}$--$99^{th}$), histogram shape (skewness, kurtosis), spatial grid means ($4 \times 4$), and per-channel color statistics, to estimate target scene brightness. Input bursts are pre-amplified during training with random jitter $\in [0.85, 1.15]$, ensuring the restoration backbone operates within a normalized latent space, invariant to the input's initial exposure.

For motion compensation, we utilize \textit{BetterSpyNet}, a coarse-to-fine optical flow estimator based on the Spatial Pyramid Network architecture \cite{ranjan2017optical}. Within the shared feature extractor, we incorporate the instance normalization~\cite{ulyanov2016instance} to achieve exposure-invariant alignment by forcing the estimator to compute flow fields from structural features rather than raw intensities. The flow is refined residually across $4$ pyramid levels: $F_i = F_{i-1}\uparrow + \Delta F_i$, where $\Delta F_i$ represents the local residual displacement computed from the concatenation of reference features, warped supporting features, and the upsampled flow $F_{i-1} \uparrow$.

To mitigate multi-frame aggregation artifacts, Temporal Attention Fusion (TAF) employs a cross-attention mechanism where reference frame features serve as the \textit{query} to evaluate the \textit{key} features of aligned supporting frames. This interaction generates spatial importance masks $M \in [0, 1]^{H \times W}$ for each frame, gating out misaligned regions, occlusions, or pixels affected by severe motion blur \cite{Dudhane23}. Weighted features are subsequently fused via $1 \times 1$ convolutions to aggregate the most reliable temporal information.

The restoration stage utilizes a modified NAFNet \cite{chen2022simple}. By replacing conventional activations with \textit{SimpleGate} operations, the architecture maintains high representational capacity with higher efficiency. Our configuration employs a base width of $32$ channels and an encoder depth of $[2, 2, 4, 8]$ with $16$ bottleneck blocks. Optimization is guided by a weighted composite loss, defined as
$\mathcal{L} = \mathcal{L}_1 + \lambda_{\text{MS-SSIM}}\mathcal{L}_{\text{MS-SSIM}} + \lambda_{\text{BC}}\mathcal{L}_{\text{BC}},$
where $\mathcal{L}_{\text{BC}}$ denotes a brightness consistency term. Final results are enhanced via 8-view test-time augmentation and post-prediction BM3D denoising \cite{dabov2009bm3d} ($\sigma \approx 1.47 \times 10^{-3}$).
\subsection{AXIOM}

Our solution improves the baseline \cite{twilightbaseline} at the components most critical for reconstruction quality: full-resolution alignment, confidence-aware burst fusion, and adaptive RAW-domain denoising. It keeps the original two-branch design with preview-based correspondence estimation and full-resolution RAW reconstruction, while strengthening alignment reliability and fusion robustness.

To make full-resolution inference practical, we optimized the matcher execution path and separated alignment/fusion from denoising into two stages, reducing peak memory usage and enabling stable deployment at target resolution. 

For burst fusion, we use frame-level reliability modeling instead of simple averaging. Each auxiliary frame is scored using matching statistics, homography inlier ratio, and reprojection error. We additionally apply Lucas--Kanade post-homography refinement \cite{lucaskanade} to reduce residual geometric errors after RANSAC. Unreliable frames are rejected, while reliable ones are fused with confidence-aware weights, improving robustness to residual misalignment, weak correspondences, and blur.

As an optional extension, low-rank adaptation \cite{lora} of the ASpanFormer \cite{chen2022aspanformer} matcher is applied to the target domain, training only 33,792 of about 15.8M parameters with pipeline-level checkpoint selection.

We further introduce local confidence weighting in the warped reference coordinate system. A residual map between each aligned auxiliary frame and the reference preview is converted into a spatial confidence map, so regions with better local agreement contribute more strongly during fusion. This reduces spatially non-uniform alignment errors and helps preserve cleaner details.

The denoising stage is performed in the RAW domain with adaptive BM3D strength estimated from the fused burst content. Unlike a fixed denoising level, this makes the method responsive to scene-dependent noise and improves reconstruction across varying low-light inputs.


\subsection{html5attention3}

\begin{figure}[t]
  \centering
   \includegraphics[width=0.6\linewidth]{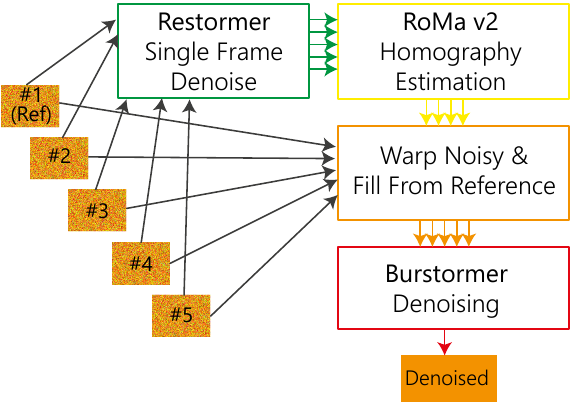}

   \caption{RestoRoBurr inference pipeline (html5attention3).}
   \label{fig:resto-ro-bur}
\end{figure}

To address the challenge we introduce RestoRoBurr (Restormer + RoMa + Burstormer) RAW-to-RAW denoising pipeline (Fig.~\ref{fig:resto-ro-bur}).

\textbf{Alignment.} Heavy misalignment in the competition images is not handled by burst denoising methods, such as Burstormer~\cite{Dudhane23}, leading to ghosting. In the meantime, warping using feature matching methods is inaccurate for highly noisy images. To mitigate the problem we (1) perform coarse alignment of the train set; (2) train Restormer~\cite{Zamir22} and denoise the training set, which exposes texture structure suppressed by shot noise, enabling accurate feature matching; (3) accurately align noisy images using their denoised counterparts via RoMav2~\cite{Edstedt25} feature matching.

\textbf{Burst Denoising.} We train Burstormer architecture using aligned images. To suppress contributions from frames with motion blur we compute full-frame Laplacian sharpness, and add Laplacian MLP gate in Burst Feature Fusion. Training objective is a combined \(\text{Charbonnier} + \gamma_1\cdot\text{Frequency} + \gamma_2\cdot\text{Gradient}\) loss, where \(\gamma_1=0.1, \gamma_2=0.2\). Image sizes are progressively increased from \(128 \times 128\) to \(256 \times 256\) during training. During inference, tiling is used with \(256 \times 256\) crops.  Training Bayer images are augmented using vertical and horizontal flips, 90 degree rotation. During second half of training hard crop mining is performed (25\% crops with the smallest loss are rejected).
\subsection{ColorWorld}

Twilight scenes exhibit extreme dynamic range where SNR varies by orders of magnitude. BadNet (6.03M params, NAFBlock-based U-Net~\cite{chen2022simple}) addresses this via three coupled mechanisms: (1)~per-scene gain pre-amplification ($g = 1/P_{99}$) normalizes input to the network's optimal representational range, where gradient flow and feature discrimination are strongest~\cite{mildenhall2018burst};
(2)~multi-frame alignment with optical flow fuses burst frames via confidence-weighted registration, avoiding ghosting artifacts; and (3)~progressive crop training ($128{\to}512$) expands receptive fields across scales, mitigating crop boundary overfitting.

At inference, tiled $512^2$ processing with AMP$\to$FP32 fallback ~\cite{Zamir22} ensures numerical stability in extreme low-SNR regions dominated by photon shot noise. The inversion $\hat{y} = f_\theta(x \cdot g) / g$ recovers the physical scale.
\subsection{mIpTMO}

We address burst RAW denoising by aligning all frames to a reference in RAW space and then denoising the averaged RAW. The pipeline avoids demosaicing before fusion so that alignment and averaging are done on the Bayer data.

\textbf{Reference and matching.} The first frame of the burst is taken as the reference. We keep both its RAW buffer and a rendered (sRGB) version. Dense correspondence is computed with RoMa~v2~\cite{Edstedt25}; matching is performed in the rendered domain to obtain a per-pixel warp and an overlap confidence map. The resulting warp and mask are then applied in RAW space so that no demosaicing artifacts are introduced before fusion.

\textbf{Alignment in RAW space.} For each remaining frame we run RoMa~v2 to get a dense warp to the reference, then transfer this warp to the RAW frame: the Bayer image is treated as four half-resolution channels, each is warped according to the same field, and the full-resolution RAW and a confidence mask are reassembled. We discard frames whose warped RAW falls below an SSIM threshold (0.90) in the overlap region relative to the reference. A weighted average of the aligned RAW frames is formed using the confidence masks.

\textbf{Denoising and output.} The averaged RAW is denoised with BM3D applied independently on the four Bayer channels. The noise level $\sigma$ is set adaptively:
$\sigma = \alpha\,\hat{\sigma} + (1 - \alpha)\,\sigma_0$, $\alpha = \alpha_0 + 0.05\,n_{\mathrm{rej}} $,
where $\hat{\sigma}$ is the noise level estimated from the normalized average via the robust median absolute deviation (MAD) of wavelet detail coefficients~\cite{donoho1994}, $\sigma_0 = 3/2047$ is a fixed base noise level, $\alpha_0 = 0.25$ is the default blending weight, and $n_{\mathrm{rej}}$ is the number of frames rejected by the SSIM filter.  When more frames are rejected, $\alpha$ grows, shifting the estimate towards $\hat{\sigma}$ to compensate for the noisier average. The BM3D~\cite{dabov2007bm3d} output is normalized (quantile stretch) and saved as a 16-bit RAW image. No learning or fine-tuning is used; we rely on pretrained RoMa~v2 for dense matching and BM3D for denoising.
\subsection{PSU}

Our proposed method operates as a two-stage pipeline: an initial multi-frame RAW fusion based on the challenge baseline, followed by a deep refinement stage.

\begin{figure}[ht]
  \centering
  \includegraphics[width=\linewidth]{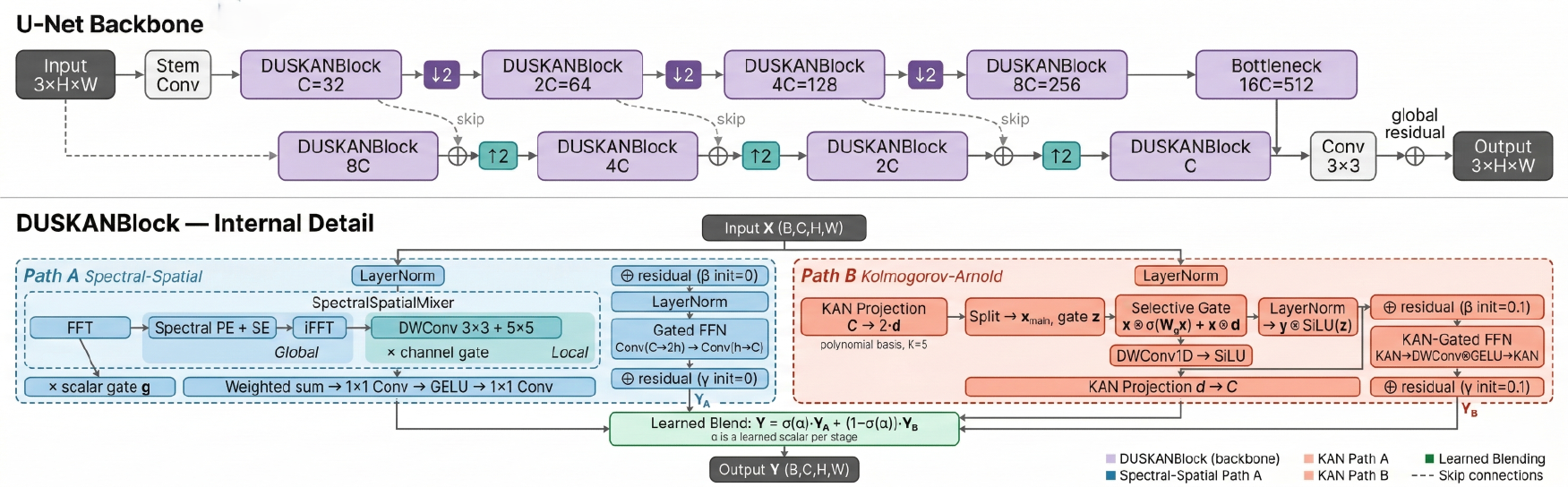}
  \caption{\textbf{DUSKAN architecture} (PSU). A learned per-stage logit $\alpha$ blends global spectral-spatial features (Path A) with polynomial-basis KAN activations (Path B) inside a symmetric 4-level U-Net.}
  \label{fig:duskan_arch}
\end{figure}

\textbf{DUSKAN Deep Refinement.}
To recover fine local texture and remove complex residual global degradations (e.g., haze, colour shifts), we introduce the \textbf{Dual Spectral Kolmogorov-Arnold Network (DUSKAN)}. The backbone is a symmetric 4-level U-Net with global residual learning. Each stage uses a DUSKANBlock (Fig.~\ref{fig:duskan_arch}), which operates two complementary branches in parallel, blended by a learned gating weight $\alpha$ to emphasise the most useful representation at each scale.

\textbf{Path A: Spectral-Spatial Processing.}
The \emph{spectral branch} computes the 2D FFT, separates magnitude from phase, and enriches the magnitude with learnable positional encodings, SE channel reweighting, and a LayerNorm-MLP refinement. The inverse-transformed global features are fused with a \emph{spatial branch} 
via a $1{\times}1$ bottleneck.

\textbf{Path B: Kolmogorov-Arnold Adaptive Processing.}
Inspired by the Kolmogorov-Arnold representation theorem, we replace fixed linear projections with a \emph{KolmogorovArnoldLinear} layer that learns custom polynomial basis activations per edge. Inside a selective gated mixer, these are combined with 1D depthwise convolutions and a parallel-additive gate to ensure stable residual gradient paths.

\textbf{Training Details.}
The U-Net uses strided $2{\times}2$ downsampling and PixelShuffle upsampling. The network is trained to minimise a combined loss function mapping pixel-level, semantic, and spectral fidelity: $\mathcal{L} = \mathcal{L}_1 + 0.01\,\mathcal{L}_{\mathrm{VGG}} + 0.1\,\mathcal{L}_{\mathrm{FF}}$.

\subsection{FengFans}

Our solution addresses multi-frame RAW denoising through a two-stage pipeline: (1)~frame alignment using SuperPoint~+~LightGlue feature matching, and (2)~learned denoising with a brightness-aware 8-channel NAFNet.

\textbf{Frame Alignment.}
We extract auto-exposed sRGB previews from each DNG file and detect SuperPoint~\cite{detone2018superpoint} keypoints (up to 2048 per frame).
LightGlue~\cite{lindenberger2023lightglue} matches keypoints between each frame and the reference. 
We estimate a homography via RANSAC and warp each Bayer sub-channel independently using scaled coordinates ($\div2$) to preserve the CFA pattern. Aligned frames are combined via masked weighted averaging.

\textbf{8-Channel Preprocessing.}
The aligned average undergoes dual-channel preprocessing:
\begin{itemize}[nosep,leftmargin=*]
  \item \textbf{Channels 1--4}: Percentile-stretched ($[p_1, p_{99}] \to [0, 1]$) Bayer average, providing normalized spatial content with consistent contrast.
  \item \textbf{Channels 5--8}: Log-compressed raw: $\log_{1001}(1 + x \times 1000)$, preserving absolute brightness.
\end{itemize}
This dual representation solves the brightness ambiguity in very dark scenes.

\textbf{NAFNet Denoiser.}
We use NAFNet~\cite{chen2022simple} with width=96, encoder blocks [2, 2, 4, 8] and 4 middle blocks (155M parameters). The model takes 8-channel input and outputs 4-channel denoised Bayer. 
At inference, we employ tiled processing (512$\times$512 tiles, 64~px overlap) and 8$\times$ TTA (4 rotations $\times$ 2 flips).

\textbf{Training.}
We employed progressive patch training with consecutive increase in patch size and Charbonnier + 0.3$\times$SSIM loss.
Stochastic Weight Averaging~\cite{izmailov2018swa} of the last 5 checkpoints from each stage was used to reduce overfitting.

\subsection{NTR}

We propose a classical pipeline for joint alignment and denoising of 5 handheld RAW Bayer frames captured under extreme low-light conditions. The pipeline operates entirely on the Bayer mosaic domain to preserve the sensor pattern.
The entire pipeline is CPU-based.

\textbf{Frame Alignment.}
We use ORB feature matching with RANSAC-based homography estimation on rendered grayscale views with frame~1 as the reference.
Rendering produces a higher-quality grayscale proxy than naive Bayer demosaicing, improving feature detection in low-light conditions.
The estimated $3\times3$ homography is applied per Bayer channel at half resolution, with a validity mask computed from the warped region.
Warping each Bayer channel separately at half resolution preserves the mosaic pattern without color aliasing artifacts.

\textbf{Multi-Frame Fusion.}
Aligned frames are fused via weighted averaging using the validity masks. 
This reduces noise while preserving spatial resolution.

Fusion is followed by BM3D denoising and quantile normalization, same as the baseline.

\section{Discussion}
\label{sec:discussion}

The NTIRE 2026 Twilight Cowboy Challenge successfully stimulated advancements in burst-based low-light enhancement, with participants achieving improvements of up to +6.49 dB in PSNR and +0.0101 in SSIM over the established baseline. 

Various strategies for robust alignment were explored, ranging from optical flow–guided deformation to transformer-based matching.
A significant trade-off between reconstruction quality and computational cost was observed. 
The winning solution by team MiAlgo leveraged extensive model ensembling and test-time augmentation to achieve state-of-the-art performance (40.62 dB PSNR), but at the cost of high inference time (514.6 s per sample).
Conversely, teams ColorWorld and DH ISP demonstrated that efficient pipelines can still deliver competitive performance, achieving mean PSNR values above 36 dB and 40 dB with inference times of 4.7 s and 10.7 s per sample, respectively. 
This suggests that while computationally intensive approaches yield gains in benchmark metrics, optimized architectures may be more suitable for practical scenarios, such as mobile deployment.
\section{Teams and affiliations}
\label{sec:teams}

\textbf{Team:} Organizers

\noindent\textbf{Members:} 
Aleksei Khalin$^{1,2,3}$,
Egor Ershov$^{1,2,3}$,
Artyom Panshin$^{1,3}$,
Sergey Korchagin$^{1,3}$,
Georgiy Lobarev$^{1,3}$,
Arseniy Terekhin$^1$,
Sofiia Dorogova$^{1,3}$,
Amir Shamsutdinov$^{1,3}$,
Yasin Mamedov$^{1,3}$,
Bakhtiyar Khalfin$^{1,3}$,
Bogdan Sheludko$^{1,3}$,
Emil Zilyaev$^{1,3}$,
Nikola Bani\'c$^4$,
Georgy Perevozchikov$^5$,
Radu Timofte$^5$

\noindent\textbf{Affiliations:}\\
$^1$Color Reproduction and Synthesis Institute, Russia\\
$^2$AXXX, Russia\\
$^3$Moscow Independent Research Institute of Artificial Intelligence, Russia\\
$^4$Gideon Brothers, Croatia\\
$^5$Computer Vision Lab, University of W\"urzburg, Germany\\

\noindent\textbf{Team:} MiAlgo

\noindent\textbf{Members:}
Shuai Liu,
Yuqian Zhang,
Lize Zhang,
Yibin Huang,
Chaoyu Feng,
Luyang Wang,
Xiaotao Wang,
Dongqing Zou,
Lei Lei

\noindent\textbf{Affiliation:} Xiaomi Inc., China\\

\noindent\textbf{Team:} DH ISP

\noindent\textbf{Members:}
Tianli Liu,
Dejun Hao,
Chunxia Lei

\noindent\textbf{Affiliation:} Zhejiang Dahua Technology Co., Ltd., China\\

\noindent\textbf{Team:} BAU-Vision

\noindent\textbf{Members:}
Furkan K\i nl\i

\noindent\textbf{Affiliation:} Department of Artificial Intelligence Engineering, Bah\c{c}e\c{s}ehir University, T\"urkiye\\

\noindent\textbf{Team:} AXIOM

\noindent\textbf{Members:}
Andrei Mironov,
Alexander Dikov,
Aleksei Sadokhin,
Vladimir Zvorygin

\noindent\textbf{Affiliation:} National Research Nuclear University MEPhI, Russia\\

\noindent\textbf{Team:} html5attention3

\noindent\textbf{Members:}
Constantine Habarlak

\noindent\textbf{Affiliation:} ConstantAI, Bulgaria\\

\noindent\textbf{Team:} ColorWorld

\noindent\textbf{Members:}
Shuwei Yue

\noindent\textbf{Affiliation:} Shenzhen Polytechnic University, China\\

\noindent\textbf{Team:} mIpTMO

\noindent\textbf{Members:}
Egor Mirantsov$^1$,
Daniil Okunev$^2$,
Dmitry Arkhipov$^2$,
Aleksandr Yugay$^1$

\noindent\textbf{Affiliations:}\\
$^1$Moscow Institute of Physics and Technology, Russia\\
$^2$ITMO University, Russia\\

\noindent\textbf{Team:} PSU

\noindent\textbf{Members:}
Anas M. Ali,
Bilel Benjdira,
Wadii Boulila

\noindent\textbf{Affiliation:} Robotics and Internet-of-Things Laboratory, Prince Sultan University, Saudi Arabia\\

\noindent\textbf{Team:} FengFans

\noindent\textbf{Members:}
Wei Zhou,
Linfeng Li,
Lingdong Kong

\noindent\textbf{Affiliation:} National University of Singapore, Singapore\\

\noindent\textbf{Team:} NTR

\noindent\textbf{Members:}
Yaoxin Jiang,
Guoyi Xu,
Jiajia Liu,
Yaokun Shi,
Jiachen Tu

\noindent\textbf{Affiliation:} University of Illinois Urbana-Champaign, United States of America
\section{Acknowledgements}
\label{sec:ackn}

This competition was carried out within the state assignment approved by the Ministry of Science and Higher Education of the Russian Federation (theme No. FFNU-2025-0045).
This work was partially supported by the Humboldt Foundation. 
We thank the NTIRE 2026 sponsors: OPPO, Kuaishou, and the University of Wurzburg (Computer Vision Lab).

{
    \small
    \bibliographystyle{ieeenat_fullname}
    \bibliography{main}
}


\end{document}